\documentclass{INTERSPEECH2023}

\interspeechcameraready

\title{Style-transfer based Speech and Audio-visual Scene understanding\\
for Robot Action Sequence Acquisition from Videos}

\name{%
  Chiori Hori${^1}$, 
  Puyuan Peng${^{1,2}}$, 
  David Harwath${^2}$, 
  Xinyu Liu${^{1,3}}$,
  Kei Ota${^1}$,\\
  \textit{
  Siddarth Jain${^1}$,
  Radu Corcodel${^1}$,
  Devesh Jha${^1}$,
  Diego Romeres${^1}$,
  Jonathan Le Roux${^1}$
  }
}
\makeatother

\address{%
${^1}$Mitsubishi Electric Research Laboratories (MERL), Cambridge, MA\\
${^2}$The University of Texas at Austin, Austin, TX \quad $^3$Brown University, Providence, RI
}
\email{\small \{chori,ota,jain,corcodel,romeres,leroux\}@merl.com}

\begin{document}
\ninept

\maketitle

\begin{abstract}
To realize human-robot collaboration, robots need to execute actions for new tasks according to human instructions given finite prior knowledge.
Human experts can share their knowledge of how to perform a task with a robot through multi-modal instructions in their demonstrations, showing a sequence of short-horizon steps to achieve a long-horizon goal.
This paper introduces a method for robot action sequence generation from instruction videos using (1) an audio-visual Transformer that converts audio-visual features and instruction speech to a sequence of robot actions called dynamic movement primitives (DMPs) and (2) style-transfer-based training that employs multi-task learning with video captioning and weakly-supervised learning with a semantic classifier to exploit unpaired video-action data. We built a system that accomplishes various cooking actions, where an arm robot executes a DMP sequence acquired from a cooking video using the audio-visual Transformer.
Experiments with Epic-Kitchen-100, YouCookII, QuerYD, and in-house instruction video datasets show that the proposed method improves the quality of DMP sequences by 
2.3 times the METEOR score 
obtained with a baseline video-to-action Transformer. The model achieved 32\% of the task success rate with the task knowledge of the object.
\end{abstract}

\section{Introduction}
\label{sec:intro}
A major goal of human-machine interaction is to develop scene-aware interaction technologies which allow machines to interact with humans based on shared knowledge obtained through recognizing and understanding their surroundings using various kinds of sensors, as introduced in \cite{IEEE_Spectrum_2022}. %
In this paper, we extend the scene-aware interaction framework to human-robot collaboration to achieve task-oriented goals.
Humans share knowledge using natural language, an abstract-level representation, and they 
can understand each other because they share similar experiences. Human students can thus achieve goals by mimicking teacher actions or manipulating target objects differently as long as they can get the same exact status as the teacher's results.
To teach human common knowledge to robots, we propose to apply scene-understanding technologies to task-oriented planning using human instruction videos, where human instructors demonstrate and explain using speech what should be done in audio-visual scenes.

To acquire human common knowledge of task oriented action sequences from human instruction videos, %
we start with the Epic-Kitchen-100 dataset~\cite{damen2022epic}, which contains egocentric cooking videos with simple short descriptions, and convert the descriptions to short-horizon steps, each of which consists of a single verb plus a few noun objects, e.g., a 5-step sequence ``{\it turn-on tap}, {\it take celery}, {\it wash celery}, {\it turn-off tap}, {\it pour celery pan},'' where the verbs and the nouns are represented with their class categories.
These action labels can be considered abstract representations for general robot actions, although a real robot of interest may not be able to perform all actions. With this dataset, we train a Transformer model that converts audio-visual features to the action sequence.

As the amount of egocentric videos in Epic-Kitchen-100 is limited 
with well-designed action labels, 
we consider using also general instruction videos from video-sharing sites such as YouTube. 
To mitigate low resource problems on the labeled data for action steps, we propose to train the model using a style-transfer approach that converts the sentence style of available video captions and subtitles (speech transcription) of a video to the action-sequence style while preserving the semantic content.
With this approach, we can generate action sequences from general instruction videos, although we still limit the video topics to ``cooking'' in this work.
For the style transfer, we apply multi-task learning and weakly-supervised learning. The multi-task learning uses action sequence generation as the primary task and video captioning as the auxiliary task, where we train two decoders for the two-style outputs on top of the shared multi-modal encoder.
The weakly-supervised learning uses a semantic classifier that judges whether the generated action sequence is semantically equivalent to the ground-truth caption sentence and uses the output as a weak label. This approach allows us to learn the decoder for action sequence generation without ground-truth action labels. Furthermore, instruction speech in a spontaneous manner can generate action sequences without audio-visual features at the inference stage as humans do on the phone without cameras.

The main contributions of this work are
(1) applying a multi-modal Transformer to generate robot actions from instruction videos,
(2) proposing a style-transfer-based approach that employs multi-task learning with video captioning and weakly-supervised learning with a semantic classifier to exploit unpaired video-action labels,
and
(3) demonstrating the effectiveness of style-transfer-based learning for robot action sequence generation in the cooking domain.

\section{Related work}
Learning robot skills from videos has been an active area of research in robotics and computer vision. At a high level, several works on robotic manipulation actions have proposed how instructions can be stored and analyzed~\cite{tenorth2013automated, yang2014cognitive, yang2015robot}. Initial work utilized contrastive learning to learn a reward function to train reinforcement learning (RL) agents~\cite{sermanet2018time}. More recently, there are some works 
using robot primitives and extraction of cost functions from task videos to enable imitation of demonstrated tasks~\cite{bahl2022human} and %
training perception modules on large sets of manipulation data to simplify learning of manipulation tasks~\cite{R3M_2022}. 
Finally, there has been growing interest in using vision and language for learning diverse robot skills. 
There are some works training visual language models using human instruction videos that are well aligned to robot actions to generate action sequences \cite{Visualtranslating4robot2018,2D/3D_RN_Robot2021}.
For example, approaches like CLIPort \cite{cliport2021} and SayCan \cite{SayCan2022} have successfully used vision-language grounding and large language models, respectively, for robot learning.
Style transfer has been applied to vision-based robot manipulation to mitigate the issues of the differences in affordance, including the kinematics of robots \cite{LinZSIL20}. 
On the other hand, our target is to convert the text style of speech instruction and video captions to a robot action sequence where those share the same context in the semantic space trained from videos.

\section{Instructional Video Action Acquisition}\label{action_detect}
This section introduces the instructional video action acquisition task, in which a system takes as input untrimmed instructional videos and outputs action sequences as verb/noun class sequences, as well as our approach to solving this problem based on dense video captioning~\cite{krishna2017dense}. %
In dense video captioning, a model needs to simultaneously segment a given long-form video into smaller clips, and caption each clip. Mathematically, given video $V$, model $f$ will be trained to produce $f(V) = \{\hat{c}_1, \hat{c}_2, \dots, \hat{c}_m\}$, where $\hat{c}_i$ is a natural language caption for segment $\hat{s}_i$ defined by onset and offset timestamps. 

\begin{figure}[tb]
    \centering
    \includegraphics[width=8.0 cm]{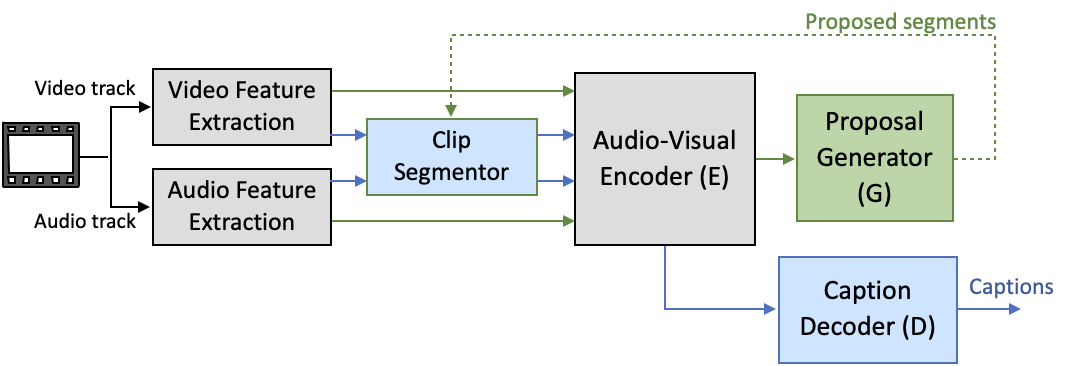}
    \vskip -2mm
    \caption{AV Transformer~\cite{iashin2020abetter} for dense-video captioning.}
    \label{fig:av_transformer}
    \vskip -5mm
\end{figure}

The model we use for this approach is a slight modification of the audio-visual (AV) Transformer model \cite{iashin2020abetter}, which contains an audio-visual encoder $E$, a caption decoder $D$, and a proposal generator $G$, as shown in Fig.~\ref{fig:av_transformer}. The audio-visual encoder has self-attention layers for each modality and cross-attention layers across modalities to better encode audio-visual features. The caption decoder is an auto-regressive Transformer decoder, which generates words by attending both to audio and visual encodings.
The proposal generator includes 1-D time-convolution modules that scan audio-visual encodings to detect segments to be captioned.
The training follows a two-stage process. The first stage is the captioning model training, where we feed the model with ground-truth video segments and train it to produce natural language captions. The second stage trains the proposal generator, where we feed the model with the entire video and train it to predict the segment timestamps. Note that the models in the two stages share the audio-visual encoder $E$.

After the two models are trained, we use the proposal generator to segment videos into clips and input the segments to the captioning model to generate captions.
In this work, we generate action-label sequences instead of natural language.
Furthermore, we skip the training of the proposal generator $G$, i.e., we use ground-truth video segments in the experiments to focus on action sequence generation. The evaluation of the complete system will be addressed in future work.

\begin{figure*}[t]
    \centering
    \vskip -1mm
    \includegraphics[width=12 cm]{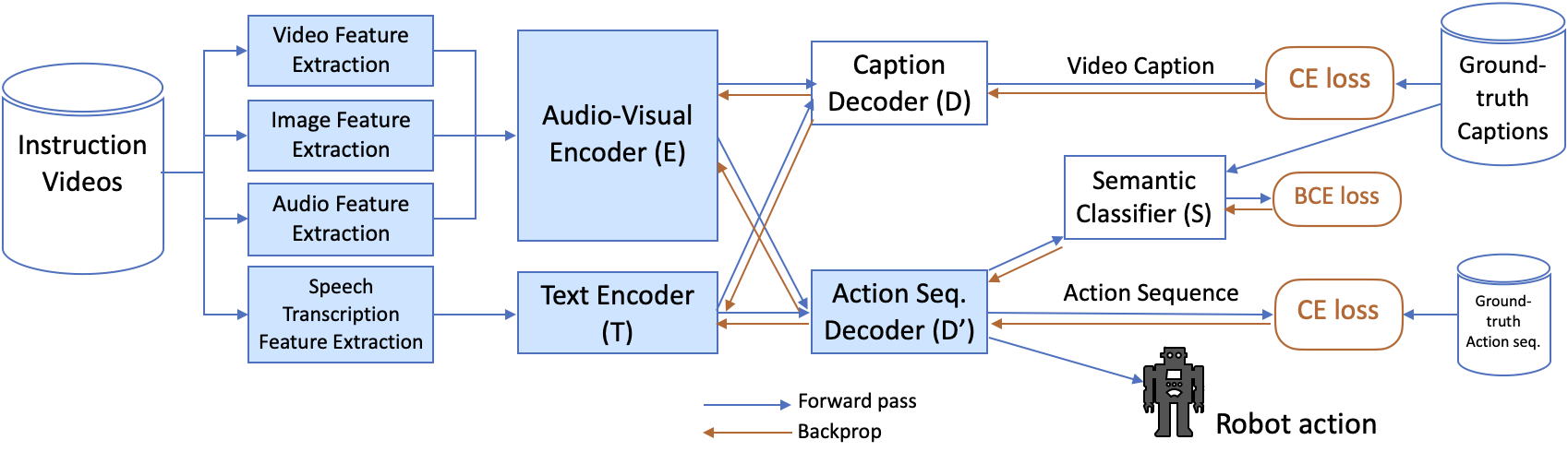}
    \vskip -2mm
    \caption{Action sequence generation model and style-transfer-based training.}
    \label{fig:action_sequence_generation}
    \vskip -4mm
\end{figure*}

\section{Action sequence generation}

\subsection{Model Architecture}
\vskip -2mm
Figure \ref{fig:action_sequence_generation} shows the action sequence generation model and additional components for training, where the model consists of the modules shaded in blue color. Given a video segment, the model generates an action sequence through the feature extraction modules, the audio-visual encoder (E), the text encoder (T), and the action sequence decoder (D'). The other modules are used at training time. 
We extend the audio-visual Transformer in Fig.~\ref{fig:av_transformer} with the text encoder, which accepts text features extracted from video subtitles. The subtitles are typically speech transcriptions provided by a speech recognizer and often include direct instructions in natural language, which potentially improves the quality of output sequences. %

\subsection{Style-transfer-based Training}
\vskip -2mm
We train the model using a style-transfer approach. %
Style transfer generally converts an image or text into different styles, but our model accepts multi-modal data including audio, video, and text (speech transcription), and generates text in different styles, i.e., action sequence and video caption, preserving the semantic content.
With this approach, we first apply multi-task learning, where the caption decoder $D$ is used for video captioning as an auxiliary task. Our aim is to acquire action sequences from general instruction videos, but the amount of instruction videos available for training is very limited since they do not have consistent action labels suitable for robots. To utilize a large number of instruction videos, we consider using video caption data that describe video scenes.
As shown in Fig.~\ref{fig:action_sequence_generation}, if the input video is annotated with an action sequence, we apply the action sequence decoder $D'$ and compute the cross-entropy (CE) loss using the ground-truth action sequence, while if the video is annotated with a caption sentence, we apply the caption decoder $D$ and compute the CE loss using the ground-truth caption.

The multi-task loss is computed as
\begin{align}
     \mathcal{L}_{\operatorname{mt}} & = \text{CE}(D(h), c) + \text{CE}(D'(h), c') \\
     h & = (E(x_A, x_V), T(x_T)),
\end{align}
where $c$ and $c'$ are the ground-truth caption sentence and action sequence, respectively.
$h$ denotes the set of audio, visual, and text encodings obtained by encoders $E$ and $T$ from corresponding feature sequences $x_A$, $x_V$, and $x_T$.
If $c$ or $c'$ does not exist for the input video, the CE loss is not computed for the missing ground truth. 
In this way, we train the shared encoders $E$ and $T$ using more data, and expect that the action-sequence-style sentences can be generated from various kinds of video recordings not limited to egocentric videos. 

We also apply weakly-supervised learning, which relies on a semantic classifier $S$ to provide weak labels.
During training, if the input video does not have an action sequence annotation but has a caption sentence, the classifier predicts whether or not the generated action sequence is semantically the same as the ground-truth caption, and we use $1$ as the weak label target. This approach allows us to train the action sequence decoder $D'$ to generate a semantically similar action sequence to the caption without ground-truth action labels.
The weakly-supervised loss is computed as
\begin{align}
     \mathcal{L}_{\operatorname{weak}} = \sum_{y' \sim D'({\bf y}|h)} \text{BCE}(S(y', c), 1), 
\end{align}
where $y'$ is sampled from the action sequence decoder $D'$ and the semantic classifier $S$ gives a probability that $y'$ and $c$ have the same semantic content. The binary cross entropy (BCE) loss is computed on the classifier output.
To perform the back-propagation, we apply continuous approximation to the decoding process \cite{yang2018unsupervised}, where we sample $y'$ from Gumbel softmax \cite{jang2016categorical} to make $y'$ differentiable. 
The classifier $S$ is separately pre-trained with positive and negative caption samples $c^+$ and $c^-$ to minimize the BCE loss
\begin{align}
   \mathcal{L}_{\operatorname{class}} = \text{BCE}(S(y, c^+), 1) + \text{BCE}(S(y, c^-), 0),
\end{align}
where we use paired captions and action sequences $(y, c^+)$ and negative samples $c^-$ randomly selected from the dataset.
When we update the generation model using $\mathcal{L}_{\operatorname{weak}}$, we freeze $S$'s parameters.

\section{Robotic Tasks and Primitive Actions}\label{robot}
This section describes the robotic tasks, the task decomposition, and our choice of representation for primitive actions. We prepared 3 common kitchen tasks %
decomposed in a sequence of subtasks: 
    (1) Make a bowl of cereal: \textit{place bowl}, \textit{pour cereal}, and \textit{pour milk}.
    (2) Make a cup of coffee: \textit{pour coffee into cup}, and \textit{pour milk into cup}.
    (3) Prepare drinks for serving: \textit{place orange juice on tray}, and \textit{place strawberry juice on tray}.

\begin{figure}[t]
     \centering
         \centering
         \includegraphics[width=0.25\textwidth]{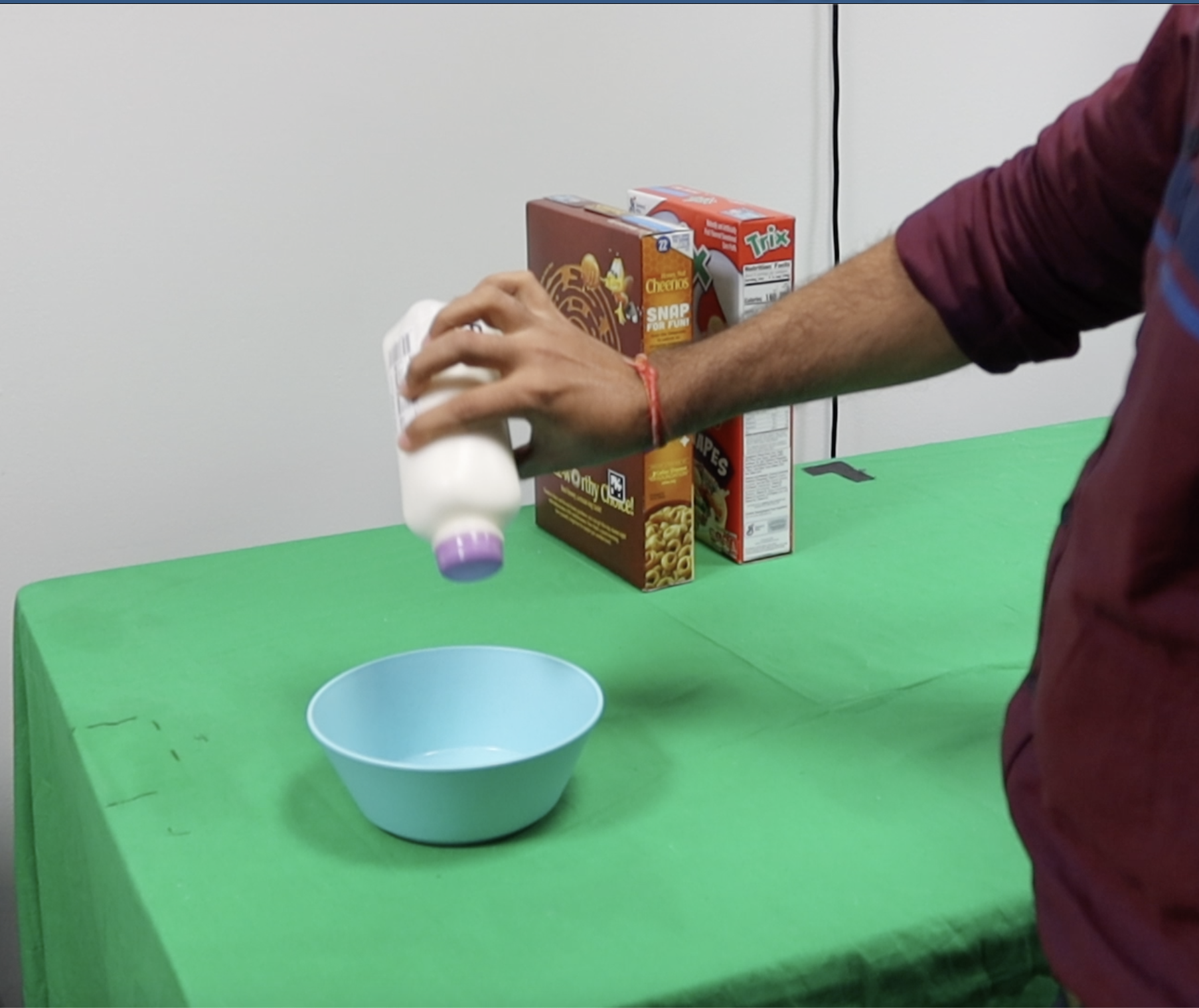}
         \vspace{-.2cm}
         \caption{Epic-Kitchen-based action labels of human instruction for ``Make a bowl of cereal'': \textit{place bowl}, \textit{pour cereal}, and \textit{pour milk}.}
         \label{fig:human}%
         \centering         \includegraphics[width=0.25\textwidth]{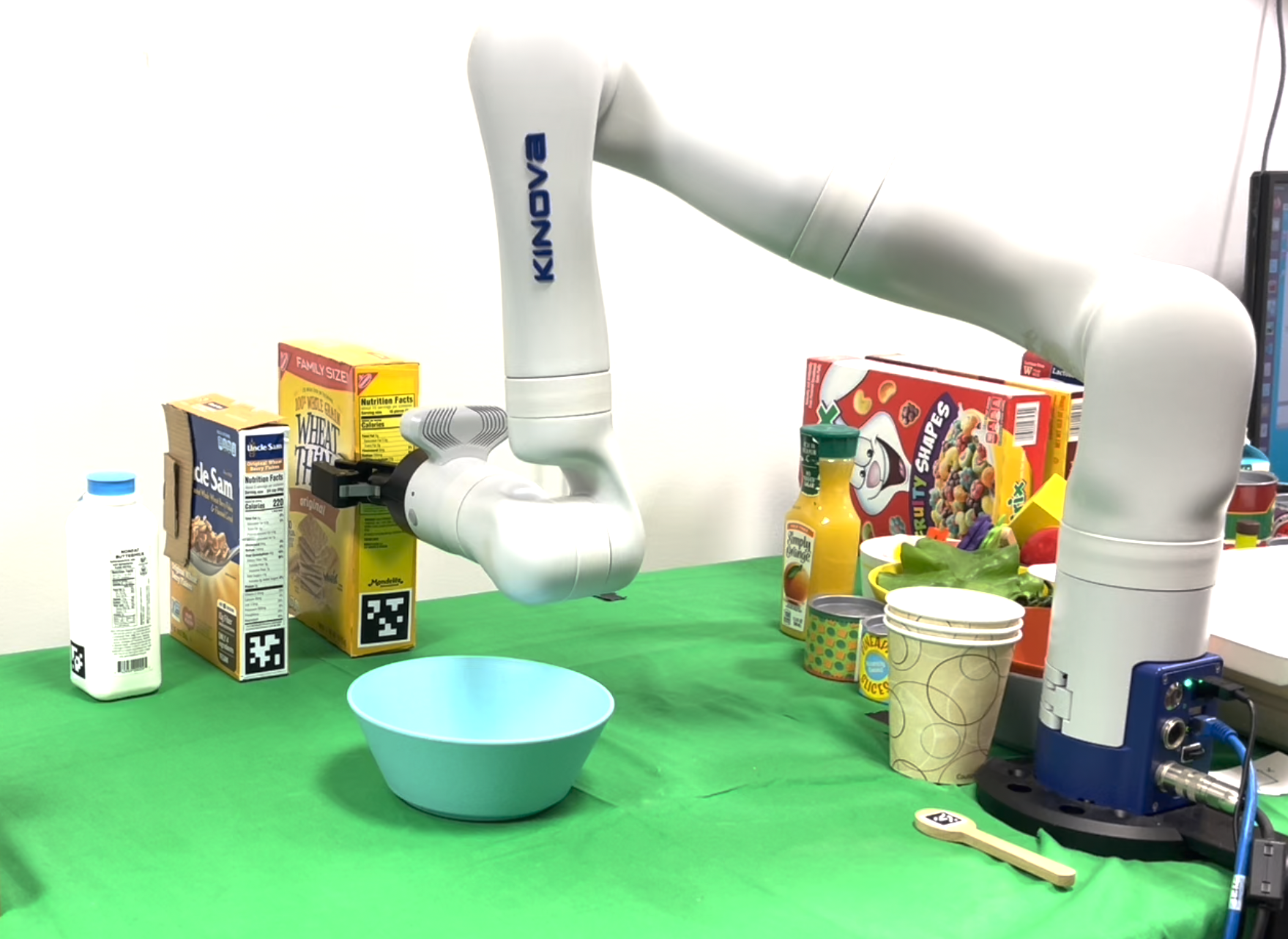}
         \vspace{-.2cm}
         \caption{DMP-based Robot action for ``Make a bowl of cereal'': \textit{top-down pick bowl}, \textit{top-down place bowl}, \textit{pick and pour cereal}, and \textit{pick and pour milk}.}
        \label{fig:robot}
     \vskip -5mm
\end{figure}

\begin{table*}[tb]
    \centering
    \small
    \caption{Video datasets. 
      ``Video [sec]'' and ``Segment [sec]'' show the average video and segment duration, respectively.
      ``Actions'', ``Captions'', and ``Subtitles'' indicate whether the videos have the corresponding annotations or not. For most videos, subtitles are speech transcriptions.
      ``(A)'' indicates testing by action generation, and ``(C)'' indicates testing by video caption generation.
    }
    \vspace{-3mm}\footnotesize
    \label{tab:datasets}
    \begin{tabular}{lcccccccccc}
        \toprule
        Dataset          & 
        \#Videos & 
        \#Segments& 
        Video [sec] & 
        Segment [sec] & 
        Actions & 
        Captions & 
        Subtitles & 
        Phase \\
        \midrule
Epic-Kitchen-100~\cite{damen2022epic} & 
        10,549 &  
        74,972 & 
        1284.2 &
        3.0 &
        \checkmark & \checkmark & & training/validation (A) \\
        YouCookII~\cite{zhou2018towards} & 
        1362   &  
        9507&
        319.1   &
        19.6     &
        &
        \checkmark & \checkmark & training/testing (A, C) \\
        QuerYD~\cite{oncescu2021queryd} (cooking) &
        88    &
        1,628    &
        192.8  &
        4.6  &
        & \checkmark & \checkmark & training/testing (C) \\
        MERL In-house  & 
        28    &  
        157  &
        22.5  &
        3.3   &
     \checkmark & \checkmark & \checkmark & validation/testing (A) \\
        \bottomrule
    \end{tabular}
    \vskip -5mm
\end{table*}

The individual tasks can be decomposed into a sequence of primitive actions.
For efficient learning, each primitive action is represented as a dynamic movement primitive (DMP) \cite{dmp,https://doi.org/10.48550/arxiv.2209.14461}. For each of these tasks, a demonstration was provided using the robot and the joystick controller for the robot. %
Each task is completed by executing a sequence of DMPs.
We use an AprilTag \cite{wang2016iros} system with an external RGB camera to detect the pose of objects during these tasks. Figure \ref{fig:robot} shows a DMP-based robot action for ``cereal.''
To generate an action sequence that could be implemented by a robot, DMPs need to be aligned to short-horizon action labels consisting of 97 verb and 300 noun classes defined by Epic-Kitchen-100 
as shown in Fig. \ref{fig:human}.
However, the DMPs need more information than the labels of Epic-Kitchen-100 in order to manipulate objects so that the robot can implement individual, object-specific tasks, such as pick an object top-down (``\textit{top-down pick}'') or place an object sideways (``\textit{side place}.'')
To align the Epic-Kitchen labels to the DMPs, we considered a set of verbs and nouns in a DMP as a ground-truth.

\section{Experiments}
\subsection{Conditions}
\vskip -2mm
We evaluate our proposed approach with instruction videos in the cooking domain from Epic-Kitchen-100~\cite{damen2022epic}, YouCookII~\cite{zhou2018towards}, QuerYD~\cite{oncescu2021queryd}, and
a newly collected in-house dataset, where Epic-Kitchen-100 consists of egocentric videos while the others consist of general cooking videos. We use a subset of QuerYD, which includes only the videos categorized into the ``cooking'' topic.
The details are summarized in Table \ref{tab:datasets}. 
We extract video features with Omnivore \cite{girdhar2022omnivore}, image features with Contrastive Language-Image Pre-Training (CLIP) \cite{radford2021learning}, and audio features with Audio Spectrogram Transformer (AST) \cite{gong21b_interspeech}. The video and image features are concatenated and projected to a single video feature sequence before feeding it to the encoder.
If a subtitle is available for a video, text features are extracted by Glove word embedding~\cite{pennington2014glove}. Otherwise, we feed an embedding vector for the \texttt{<unk>} label instead.
The numbers of dimensions of audio, visual, and text features are 768, 1024, and 300, respectively.\\
\indent The audio-visual Transformer contains audio-visual and text encoders with two-layer blocks, where the dimensions of multi-head attentions are $d_{\it model}^{(V)}=d_{\it model}^{(A)}=768$ for audio-visual layer blocks and $d_{\it model}^{(T)}=300$ for text encoder.
The dimensions of the feed-forward layers are set as $d_{\it ff}^{(*)}=4 \times d_{\it model}^{(*)}$. The action sequence decoder consists of two-layer blocks, where $d_{\it model}^{(D)}=300$. The caption decoder has the same architecture as the action sequence decoder.
The number of attention heads is 4 for all the Transformer layer blocks. The semantic classifier is a two-layer feed-forward network that accepts two text feature vectors after mean pooling over each word embedding vector sequence and outputs a probability that the input vectors have the same semantic content. The number of dimensions of the hidden layer is 300 and the output dimension is 1, which is converted to a probability by the sigmoid function.\\
\indent
To evaluate the quality of generated action sequences, we use BLEU-1, BLEU-2, and METEOR scores computed between the generated and ground-truth sequences as used in the robotics field \cite{Visualtranslating4robot2018,2D/3D_RN_Robot2021}. 
Additionaly, the task success rate was evaluated.
Since the size of the in-house dataset is small, we conduct a 3-fold cross validation using 
28 videos 
in the dataset, where we split the dataset into 3 subsets consisting of 9, 9, and 10 videos and use each subset for testing and the rest for validation. Each validation set is used to choose the best epoch model based on the METEOR score. In this paper, we report the average scores over the three subsets.

\subsection{Results}
\vskip -2mm
Table~\ref{tab:result} shows the quality of generated action sequences using different models, where ``Baseline'' denotes the model trained with only the Epic-Kitchen-100 dataset without the caption decoder. ``Multi-task'' indicates that the model was trained by multi-task learning together with the caption decoder using all the datasets for training. ``+Weak-sup.'' indicates that we fine-tuned the model with weakly-supervised learning after multi-task learning. In the fine-tuning process, we used the sum of multi-task loss $\mathcal{L}_{\operatorname{mt}}$ and weakly-supervised loss $\mathcal{L}_{\operatorname{weak}}$.

The baseline model provided high BLEU/METEOR scores for the EpicKitchen-100 validation set, while it did not perform well for the in-house dataset due to %
the mismatches in video recording conditions and instruction styles, e.g., with ego-centric or distant camera, and with narration, music, or silence.
For the in-house data, we obtained substantial improvement over the baseline using multi-task learning. For example, the METEOR score is 2.15 times that obtained with baseline.
With weakly-supervised learning, 
we further obtained additional 0.16 times and a total improvement from the baseline is 2.3 times the METEOR score of the baseline.
Additionally, we tested the impact of the ASR errors using Google API.
The ASR results with 30.9 WER slightly degraded the performance.
Additionally, we tested the development sets of YouCookII. The variation of the scenarios is broader; thus, 
the total improvement is 1.4 times the baseline in METEOR, which is worse than that of the in-house data.
Supplementally note that the caption decoder trained in a multi-task training manner achieved 0.17 and 0.06 in METEOR using a single reference on YouCookII and QuerYD, respectively. The scores were almost comparable with those for video captioning solo decoder trained using YouCookII and QuerYD. This shows the caption decoder works reasonably and supports semantic representation for action sequence generation.
\begin{table}[tbh]
    \centering
    \small
    \caption{Generated action sequence quality. 
    YouCookII test set has 50 videos with 1513 actions.
    The row with * shows the impact of the ASR results with 30.9 WER by Google API.}
    \vskip -2mm
    \footnotesize
    \label{tab:result}
    \begin{tabular}{lcccc}
        \toprule
                     & Eval. set & BLEU-1 & BLEU-2 & METEOR \\
        \midrule
        
        Baseline	&	Epic-Kitchen	&	0.499	&	0.374	&	0.296	\\
        Baseline	&	In-house	&	0.228	&	0.049	& 	0.101	\\
        Multi-task	&	In-house	&	0.402	&	0.254	&	0.217	\\
        +Weak-sup.	&	In-house	&	\bf 0.418	&	\bf 0.273	&	\bf 0.233	\\
    +Weak-sup.$^*$ &	In-house	&	0.414	&	0.266	&	0.228	\\
        \midrule
        Baseline	&	YouCookII &	 0.160	&	0.022	&	0.072	\\
        Multi-task	&	YouCookII &	 0.215	&	0.079	&	0.096	\\
        +Weak-sup   &	YouCookII &	 \bf 0.227	&	\bf 0.085	&	\bf 0.104	\\	
        \bottomrule
    \end{tabular}
    \\ 
    \vskip -3mm
\end{table}

\begin{table}[tbh]
    \centering
    \small
    \caption{Ablation result. Each row shows the result when removing the indicated feature during training and/or testing. }%
    \vskip -2mm
    \footnotesize
    \label{tab:ablation}
    \begin{tabular}{ccccc}
        \toprule
        Training   & Testing   & BLEU-1 & BLEU-2 & METEOR \\
        
        \midrule
	-		    & -		        &   \bf 0.418	&	\bf 0.273	&	\bf 0.233	\\
        audio		& audio		    &	0.350	&	0.196	&	0.172	\\	
	-		    & audio		    &	0.405	&	0.258	&	0.227	\\
	-		    & subtitle		&	0.411	&	0.252	&	0.227	\\
        subtitle	& subtitle		&	0.382	&	0.232	&	0.202	\\
	-		    & video/image   &	0.126	&	0.072	&	0.064	\\
        \bottomrule
    \end{tabular}
    \vskip -5mm
\end{table}

Table \ref{tab:ablation} shows the result of an ablation study, where we removed specific features from training and/or testing. The top row corresponds to our best system without any ablations. 
The audio features represent a wide variety of audio information, including speech, event sounds, and noise. 
The model trained w/o the audio features degraded the performance. 
This implies the audio feature characterizes the scenes weakly as shown in \cite{hori2017attention}
. 
The subtitle %
features are essential for training. 
The degradation in the last row w/o using video/image features implies the data is not sufficient to train a model characterizing actions using only speech instruction. 

Table \ref{tab:success} shows
the task success rate under the assumption that the task can be successfully completed if all actions in the video clip are predicted correctly. 
With task knowledge, we masked out unrelated objects which do not exist on the workbench for the robot from the target vocabulary in micro-step generation.  
\begin{table}[h]
    \centering
    \vskip -3mm
    \caption{Task success evaluation }
    \footnotesize
    \vskip -3mm
    \label{tab:success}
    \footnotesize
    \begin{tabular}{cccc}
        \toprule
        Task    & Word error & Action error & Task success \\
        knowledge   & \lbrack \%\rbrack & \lbrack \%\rbrack & rate \lbrack \%\rbrack \\
        \midrule
        
	            &	56.8		&		82.2	&	10.7		\\
        \checkmark &	36.6 		&		55.4	&	32.1	\\
        \bottomrule
    \end{tabular}
    \vskip -5mm
\end{table}

\section{Conclusions}
This paper proposed a method for generating robot action sequences from instruction videos. We use an audio-visual Transformer that converts audio-visual features and instruction speech to a sequence of robot actions. Additionally, we utilize style-transfer-based training that employs multi-task learning with video captioning and weakly-supervised learning with a semantic classifier to exploit unpaired video-action data. 
Experiments with instruction videos from Epic-Kitchen-100, YouCookII, QuerYD, and a newly collected in-house dataset demonstrated that our proposed method improves the quality of action sequences 
 by 2.3 times the METEOR score obtained with a baseline video-to-action Transformer. 
The best model achieved 32\% in task success rate with the task knowledge.

\bibliographystyle{IEEEtran}
\bibliography{mybib}

\begin{thebibliography}{10}
\providecommand{\url}[1]{#1}
\csname url@samestyle\endcsname
\providecommand{\newblock}{\relax}
\providecommand{\bibinfo}[2]{#2}
\providecommand{\BIBentrySTDinterwordspacing}{\spaceskip=0pt\relax}
\providecommand{\BIBentryALTinterwordstretchfactor}{4}
\providecommand{\BIBentryALTinterwordspacing}{\spaceskip=\fontdimen2\font plus
\BIBentryALTinterwordstretchfactor\fontdimen3\font minus
  \fontdimen4\font\relax}
\providecommand{\BIBforeignlanguage}[2]{{%
\expandafter\ifx\csname l@#1\endcsname\relax
\typeout{** WARNING: IEEEtran.bst: No hyphenation pattern has been}%
\typeout{** loaded for the language `#1'. Using the pattern for}%
\typeout{** the default language instead.}%
\else
\language=\csname l@#1\endcsname
\fi
#2}}
\providecommand{\BIBdecl}{\relax}
\BIBdecl

\bibitem{IEEE_Spectrum_2022}
C.~Hori and A.~Vetro, ``At last, a self-driving car that can explain itself,''
  \emph{IEEE Spectrum}, Feb. 2022.

\bibitem{damen2022epic}
D.~Damen, H.~Doughty, G.~M. Farinella, A.~Furnari, E.~Kazakos, J.~Ma,
  D.~Moltisanti, J.~Munro, T.~Perrett, W.~Price \emph{et~al.},
  ``Epic-kitchens-100,'' \emph{International Journal of Computer Vision}, vol.
  130, pp. 33--55, 2022.

\bibitem{tenorth2013automated}
M.~Tenorth, J.~Ziegltrum, and M.~Beetz, ``Automated alignment of specifications
  of everyday manipulation tasks,'' in \emph{2013 IEEE/RSJ International
  Conference on Intelligent Robots and Systems}, 2013, pp. 5923--5928.

\bibitem{yang2014cognitive}
Y.~Yang, A.~Guha, C.~Fermuller, and Y.~Aloimonos, ``A cognitive system for
  understanding human manipulation actions,'' \emph{Advances in Cognitive
  Sysytems}, vol.~3, pp. 67--86, 2014.

\bibitem{yang2015robot}
Y.~Yang, Y.~Li, C.~Fermuller, and Y.~Aloimonos, ``Robot learning manipulation
  action plans by" watching" unconstrained videos from the world wide web,'' in
  \emph{Proceedings of the AAAI Conference on Artificial Intelligence},
  vol.~29, no.~1, 2015.

\bibitem{sermanet2018time}
P.~Sermanet, C.~Lynch, Y.~Chebotar, J.~Hsu, E.~Jang, S.~Schaal, S.~Levine, and
  G.~Brain, ``Time-contrastive networks: Self-supervised learning from video,''
  in \emph{2018 IEEE international conference on robotics and automation
  (ICRA)}.\hskip 1em plus 0.5em minus 0.4em\relax IEEE, 2018, pp. 1134--1141.

\bibitem{bahl2022human}
S.~Bahl, A.~Gupta, and D.~Pathak, ``Human-to-robot imitation in the wild,'' in
  \emph{Proc. RSS}, 2022.

\bibitem{R3M_2022}
S.~Nair, A.~Rajeswaran, V.~Kumar, C.~Finn, and A.~Gupta, ``{R3M}: A universal
  visual representation for robot manipulation,'' \emph{arXiv preprint
  arXiv:2203.12601}, 2022.

\bibitem{Visualtranslating4robot2018}
A.~Nguyen, D.~Kanoulas, L.~Muratore, D.~G. Caldwell, and N.~G. Tsagarakis,
  ``Translating videos to commands for robotic manipulation with deep recurrent
  neural networks,'' in \emph{2018 IEEE International Conference on Robotics
  and Automation (ICRA)}, 2018, pp. 3782--3788.

\bibitem{2D/3D_RN_Robot2021}
X.~Xu, K.~Qian, B.~Zhou, S.~Chen, and Y.~Li, ``Two-stream 2d/3d residual
  networks for learning robot manipulations from human demonstration videos,''
  in \emph{2021 IEEE International Conference on Robotics and Automation
  (ICRA)}, 2021, pp. 3353--3358.

\bibitem{cliport2021}
M.~Shridhar, L.~Manuelli, and D.~Fox, ``Cliport: What and where pathways for
  robotic manipulation,'' in \emph{Proc. CoRL}, 2021.

\bibitem{SayCan2022}
M.~Ahn, A.~Brohan, N.~Brown, Y.~Chebotar, O.~Cortes, B.~David, C.~Finn, C.~Fu,
  K.~Gopalakrishnan, K.~Hausman \emph{et~al.}, ``Do as {I} can, not as {I} say:
  Grounding language in robotic affordances,'' \emph{arXiv preprint
  arXiv:2204.01691}, 2022.

\bibitem{LinZSIL20}
Y.-C. Lin, A.~Zeng, S.~Song, P.~Isola, and T.-Y. Lin, ``Learning to see before
  learning to act: Visual pre-training for manipulation,'' in \emph{Proc.
  ICRA}, 2020, pp. 7286--7293.

\bibitem{krishna2017dense}
R.~Krishna, K.~Hata, F.~Ren, L.~Fei-Fei, and J.~C. Niebles, ``Dense-captioning
  events in videos,'' in \emph{Proc. ICCV}, Oct. 2017, pp. 706--715.

\bibitem{iashin2020abetter}
V.~Iashin and E.~Rahtu, ``A better use of audio-visual cues: Dense video
  captioning with bi-modal transformer,'' in \emph{Proc. BMVC}, 2020.

\bibitem{yang2018unsupervised}
Z.~Yang, Z.~Hu, C.~Dyer, E.~P. Xing, and T.~Berg-Kirkpatrick, ``Unsupervised
  text style transfer using language models as discriminators,'' in \emph{Proc.
  NeurIPS}, 2018.

\bibitem{jang2016categorical}
E.~Jang, S.~Gu, and B.~Poole, ``Categorical reparameterization with
  gumbel-softmax,'' \emph{arXiv preprint arXiv:1611.01144}, 2016.

\bibitem{zhou2018towards}
L.~Zhou, C.~Xu, and J.~J. Corso, ``Towards automatic learning of procedures
  from web instructional videos,'' in \emph{Proc. AAAI}, 2018.

\bibitem{oncescu2021queryd}
A.-M. Oncescu, J.~F. Henriques, Y.~Liu, A.~Zisserman, and S.~Albanie,
  ``Quer{YD}: A video dataset with high-quality text and audio narrations,'' in
  \emph{Proc. ICASSP}, 2021, pp. 2265--2269.

\bibitem{dmp}
M.~Saveriano, F.~J. Abu{-}Dakka, A.~Kramberger, and L.~Peternel, ``Dynamic
  movement primitives in robotics: {A} tutorial survey,'' \emph{arXiv preprint
  arXiv:2102.03861}, 2021.

\bibitem{https://doi.org/10.48550/arxiv.2209.14461}
\BIBentryALTinterwordspacing
S.~Shaw, D.~K. Jha, A.~Raghunathan, R.~Corcodel, D.~Romeres, G.~Konidaris, and
  D.~Nikovski, ``Constrained dynamic movement primitives for safe learning of
  motor skills,'' 2022. [Online]. Available:
  \url{https://arxiv.org/abs/2209.14461}
\BIBentrySTDinterwordspacing

\bibitem{wang2016iros}
J.~Wang and E.~Olson, ``{AprilTag} 2: Efficient and robust fiducial
  detection,'' in \emph{Proc. IROS}, Oct. 2016.

\bibitem{girdhar2022omnivore}
R.~Girdhar, M.~Singh, N.~Ravi, L.~van~der Maaten, A.~Joulin, and I.~Misra,
  ``{Omnivore: A Single Model for Many Visual Modalities},'' in \emph{Proc.
  CVPR}, 2022.

\bibitem{radford2021learning}
A.~Radford, J.~W. Kim, C.~Hallacy, A.~Ramesh, G.~Goh, S.~Agarwal, G.~Sastry,
  A.~Askell, P.~Mishkin, J.~Clark \emph{et~al.}, ``Learning transferable visual
  models from natural language supervision,'' in \emph{Proc. ICML}, 2021, pp.
  8748--8763.

\bibitem{gong21b_interspeech}
Y.~Gong, Y.-A. Chung, and J.~Glass, ``{AST: Audio Spectrogram Transformer},''
  in \emph{Proc. Interspeech}, 2021, pp. 571--575.

\bibitem{pennington2014glove}
J.~Pennington, R.~Socher, and C.~D. Manning, ``Glove: Global vectors for word
  representation,'' in \emph{Proc. EMNLP}, 2014, pp. 1532--1543.

\bibitem{hori2017attention}
C.~Hori, T.~Hori, T.-Y. Lee, Z.~Zhang, B.~Harsham, J.~R. Hershey, T.~K. Marks,
  and K.~Sumi, ``Attention-based multimodal fusion for video description,'' in
  \emph{Proc. ICCV}, 2017.

\end{thebibliography}

\end{document}